\documentclass{article} 
\usepackage{conference,times}

\usepackage{amsmath,amsfonts,bm}

\def\eqref#1{equation~\ref{#1}}

\def\1{\bm{1}}

\DeclareMathAlphabet{\mathsfit}{\encodingdefault}{\sfdefault}{m}{sl}
\SetMathAlphabet{\mathsfit}{bold}{\encodingdefault}{\sfdefault}{bx}{n}

\usepackage{algorithm}
\usepackage{algpseudocode}  
\usepackage{hyperref}
\usepackage{url}
\usepackage{multirow}
\usepackage{booktabs}
\usepackage{caption}
\usepackage{graphicx}
\usepackage{wrapfig}
\usepackage{tikz}
\usetikzlibrary{arrows.meta} 
\usepackage{enumitem}

\newif\ifanon
\anonfalse

\title{Flow-Matching Based Refiner for Molecular Conformer Generation}
\ifanon
  \author{Anonymous Authors}
\else
  \author{
    Xiangyang Xu \quad Hongyang Gao  \\
    Iowa State University \\
    \texttt{xyxu@iastate.edu, hygao@iastate.edu}
  }
\fi

\begin{document}
\maketitle

\begin{abstract}
Low-energy molecular conformers generation (MCG) is a foundational yet challenging problem in drug discovery. 
Denoising-based methods include diffusion and flow-matching methods that learn mappings from a simple base distribution to the molecular conformer distribution. 
However, these approaches often suffer from error accumulation during sampling, especially in the low SNR steps, which are hard to train. 
To address these challenges, we propose a flow-matching refiner for the MCG task. 
The proposed method initializes sampling from mixed-quality outputs produced by upstream denoising models and reschedules the noise scale to bypass the low-SNR phase, thereby improving sample quality. 
On the GEOM-QM9 and GEOM-Drugs benchmark datasets, the generator–refiner pipeline improves quality with fewer total denoising steps while preserving diversity. 
\end{abstract}


\section{Introduction}

Low-energy 3D conformations, called \emph{conformers}, determine a molecule’s biological, 
chemical, and physical properties. 
\citep{ guimaraes2012, schutt2018schnet, klicpera2019directional, axelrod2020molecular, schutt2021equivariant}
Therefore, generating accurate and diverse ensembles of conformers from the molecular graph is a fundamental task in computational chemistry.
Traditional approaches can be grouped into two main categories: physics-based methods, such as molecular dynamics\citep{bkgd_md_pracht2020automated}, which explore conformational space with high fidelity at high computational cost; and cheminformatics methods \citep{bkgd_cheminfo_hawkins2010conformer, bkgd_cheminfo_riniker2015better}, which are more efficient but less accurate, often trading precision for speed.

Denoising-based generative models, including diffusion \citep{diff_ho2020denoising, diff_song2020score} and flow matching \citep{fm_lipman2022flow, fm_albergo2022building, fm_liu2022flow, fm_tong2023improving}, have developed rapidly in recent years. Recent machine learning methods address this gap in the MCG task by learning to sample the distribution of low-energy conformers, aiming for both speed and accuracy. Existing work includes (1) methods based on pairwise atom distances in a distance matrix \citep{bkgd_dm_luo2021predicting, bkgd_dm_shi2021learning}; however, they have too many degrees of freedom, which often leads to unstable optimization and poor results. (2) methods operate in torsional angle space \citep{baseline_jing2022torsional}; unfortunately, the performance is limited by the use of local information obtained by preprocessing tools such as RDKit. (3) More recent methods that predict 3D coordinates and report the current SOTA performance \citep{baseline_xu2022geodiff, baseline_wang2024swallowing, baseline_hassan2024flow, baseline_liu2025next}.  They work directly in 3D coordinate space to achieve the current best results, especially with large-scale models. However, the methods based on 3D atom coordinates still suffer issues from the denosing method: models are hard to train to predict the score or vector field of steps dominated by noise. \citet{motivation_karras2022elucidating}. Early diffusion steps that start from pure noise incur large errors. Because denoising is sequential and removes noise gradually, these early errors can accumulate and propagate through the trajectory, degrading final quality \citep{motivation_LiS24a, err-acc-li2023error, err-acc-li2023q, err-acc-chung2022improving}.

\begin{figure*}[ht]
  \centering
  \includegraphics[width=0.9\linewidth,clip,trim=0 200 0 75]{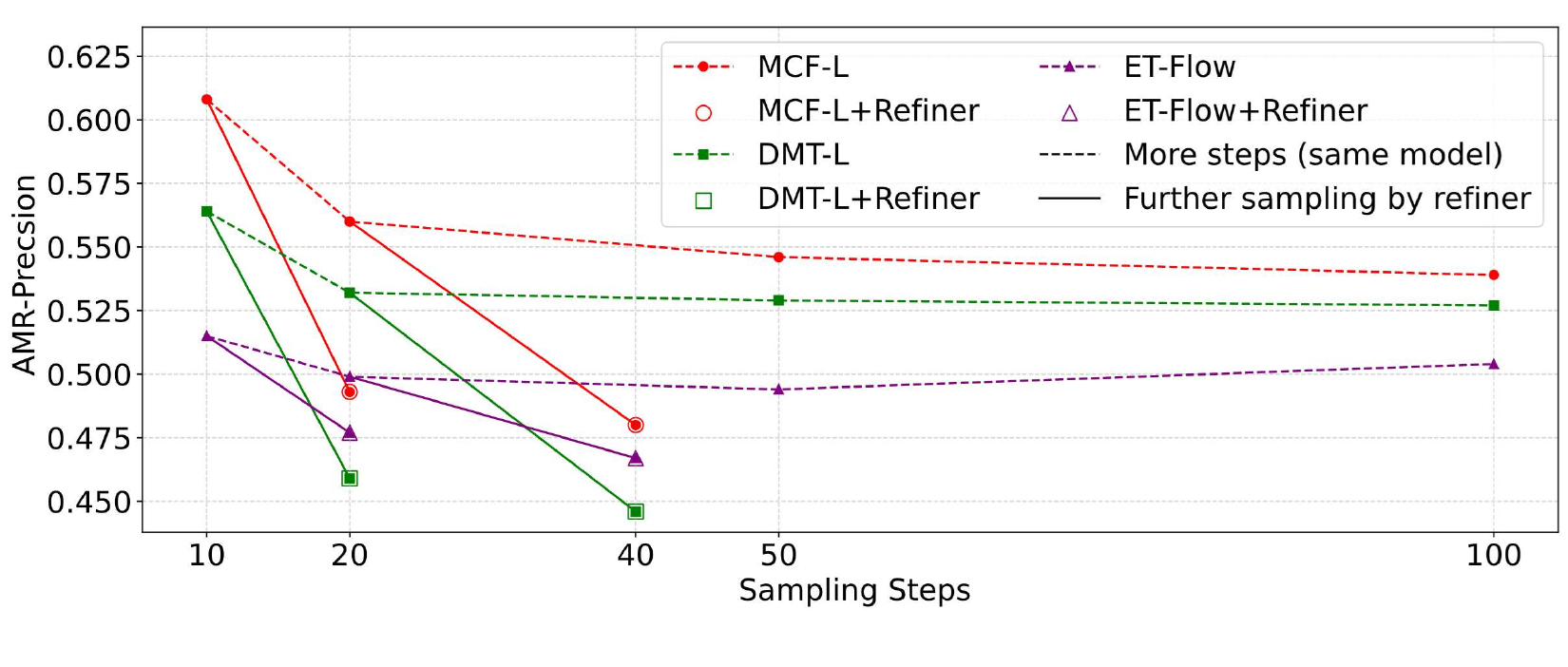}
\caption{GEOM-DRUGS: increasing sampling steps vs.\ adding a refiner.
Metric: Average Minimum RMSD (AMR)$\downarrow$.
The x-axis shows \emph{total} sampling steps (including refiner steps).
We compare allocating extra budget to adding our refiner (8.3M) against simply increasing steps for three strong baselines: MCF-L (242M), ET-Flow (8.3M), and DMT-L (150M).
For ET-Flow, the base architecture and predictor--corrector (PC) schedule are identical; only the refiner is added and fine-tuned.}
  \label{fig:quality-vs-steps}
\end{figure*}

We address this by coupling a standard denoising model with a refiner. Instead of beginning from pure noise, the refiner starts from an upstream generator’s conformer: a plausible structure perturbed with error, which can be considered as some noise; rather than a noise-only state. By skipping the high-noise phase, early-step errors are substantially smaller than in traditional denoising, improving final performance.

Empirically, as shown in Fig.~\ref{fig:quality-vs-steps}, simply adding steps improves only marginally, whereas our method achieves larger improvements with fewer additional steps; this even holds under a fixed model architecture with the same PC sampler.

Our contributions can be summarized as follows:
\begin{enumerate}
\item We introduce a denoising model plus \emph{refiner} pipeline for molecular conformer generation and theoretically justify its design.
\item We identify the properties that make the refiner effective and validate them theoretically and empirically.
\item The pipeline reaches higher quality with fewer total sampling steps; this gain remains even when the refiner model structure is the same as the previous method.
\item The proposed refiner is plug-and-play across the conformer generated by different upstream models and needs no per-model tuning.
\end{enumerate}

\section{Related Work}
\textbf{Diffusion and Flow Matching.}
Diffusion models \citep{diff_song2020score, diff_ho2020denoising} generate high-quality, diverse samples by learning the score $\nabla_{\mathbf x}\log p_t(\mathbf x)$ of noise-perturbed data and numerically integrating a reverse-time SDE/ODE from a Gaussian prior to the data distribution.
Flow Matching (FM) \citep{fm_lipman2022flow, fm_albergo2022building,  fm_liu2022flow} instead regresses a target vector field defined by stochastic interpolants, which also enables training with arbitrary source distributions  \citep{fm_pooladian2023multisample}.
Recent work unifies diffusion and FM under stochastic interpolants, revealing them as closely related denoising generative models. \citep{df-unify-albergo2023stochastic, df-unify-ma2024sit, df-unify-yu2024representation}.
However, in very low SNR regimes where noise dominates, the training targets become hard to learn \citep{motivation_karras2022elucidating, bkdg-uloss-hang2024improved}, and during the early denoising steps, errors can accumulate and propagate, degrading sample quality \citep{motivation_LiS24a,err-acc-chung2022improving,err-acc-li2023error,err-acc-li2023q},.


\textbf{Refiners.}
Refinement modules are widely used in coarse-to-fine pipelines to recover high-frequency detail. \citet{bkgd-refiner-podell2023sdxl} introduces a diffusion-time–conditioned stage that polishes images, and \citet{bkgd-refiner-pandey2022diffusevae} refines reconstructions produced by an upstream VAE.
Restoration-style refiners target specific degradations such as deblurring and super-resolution \citep{bkgd-refiner-whang2022deblurring,bkgd-refiner-saharia2022image}, However, and typically rely on a particular upstream model or error type.
Other methods repair outputs by injecting noise and then denoising \citep{bkgd-refiner-kawar2022denoising,bkgd-refiner-sawata2022diffiner}; however, such noise-injection–based refiners must still start sampling by revisiting high-noise steps thereby risking instability.

\textbf{Molecular Conformer Generation.}
Deep learning methods for molecular conformer generation have attracted growing interest in recent years.
Early attempts applied diffusion to distance matrices \citep{bkgd_dm_luo2021predicting, bkgd_dm_shi2021learning}, but these approaches underperformed.
\citet{baseline_ganea2021geomol} predict conformers via learned chemical structure parameters (e.g., bond lengths, bond angles, and torsions), yet overall quality is constrained by torsional accuracy; \citet{baseline_jing2022torsional} improves torsion prediction but remains limited by local structure.
Subsequent works model and perturb directly in atomic coordinate space and scale to larger architectures \citep{baseline_xu2022geodiff, baseline_wang2024swallowing, baseline_hassan2024flow, baseline_liu2025next}, achieving state-of-the-art performance.
Nevertheless, sampling quality remains constrained by the intrinsic behavior of sequential denoising procedures.
\section{Preliminaries}

\textbf{Flow Matching for 3D Molecular Conformers}: Following \citet{baseline_hassan2024flow}, Given a molecular graph \(\mathcal G\) and all-atom coordinates \(\mathbf x\in\mathbb{R}^{N\times 3}\),
flow matching learns a time-dependent \(\mathrm{SE}(3)\)-equivariant vector field
\(\mathbf v_\theta(\mathbf x,t, \mathcal G)\) that transports a tractable base
to the low-energy target.

\textbf{Interpolant and training.}
Sample a coupling \((\mathbf x_0,\mathbf x_1)\sim\rho_0\times\rho_1\) (conditioned on \(\mathcal G\)),
and define a interpolant with stochastic 
\begin{equation}
\label{eq:interp}
\mathbf x_t \;=\; \alpha(t)\,\mathbf x_0 \;+\; \beta(t)\,\mathbf x_1 \;+\; s(t)\,\mathbf z,
\qquad t\!\in\![0,1],\ \ \mathbf z\!\sim\!\mathcal N(\mathbf 0,\mathbf I),
\end{equation}
with \(\alpha(0)=1,\ \beta(0)=0,\ \alpha(1)=0,\ \beta(1)=1\).
Its instantaneous (non-parametric) velocity is
\begin{equation}
\label{eq:u}
\mathbf u_t \;:=\; \tfrac{\mathrm d}{\mathrm dt}\mathbf x_t
\;=\; \alpha'(t)\,\mathbf x_0 \;+\; \beta'(t)\,\mathbf x_1 \;+\; s'(t)\,\mathbf z .
\end{equation}
The optimal transport velocity is
\(\mathbf v_t^\star(\mathbf x)=\mathbb E[\mathbf u_t\mid \mathbf x_t=\mathbf x,\,t]\).
Flow matching trains the model \(\mathbf v_\theta\) via supervised regression:
\begin{equation}
\label{eq:fm-loss-preliminary}
\mathcal L_{\text{FM}}(\theta)=
\mathbb E\Big[\big\|\,\mathbf v_\theta(\mathbf x_t,t,G)-\mathbf u_t\,\big\|_2^2\Big],
\end{equation}

The Base priors \(\mathbf x_0\) is: a bonded harmonic prior is used to preserve local geometry, or pure Gaussian noise: \(\mathbf x_0=\sigma\,\boldsymbol{\varepsilon}\). Therefore, when training and starting sampling at $t=0$, there's no signal.


\section{Flow-Matching-Based Refiner}\label{sec: method}

\textbf{Motivation.}
Current molecular conformer generation (MCG) methods typically rely on denoising generative models (diffusion or flow matching), which currently achieve state-of-the-art performance \citep{baseline_hassan2024flow, baseline_liu2025next, baseline_wang2024swallowing} compared to alternative paradigms.
However, early high-noise sampling steps are difficult to learn  \citep{motivation_karras2022elucidating}; when sampling starts from pure noise, the initial stage suffers from large errors.
At the same time, denoising sampling is sequential; these stepwise errors will propagate and accumulate along the sampling process, and finally harm the final quality. \citep{motivation_LiS24a, err-acc-chung2022improving, err-acc-li2023error, err-acc-li2023q}

For the MCG task, we work in $\mathbf x \in \mathbb{R}^{3N}$ for a molecular conformer with $N$ atoms. Let $\mathbf{x}_1$ be the ground-truth coordinates and $\hat{\mathbf{x}_1}$ denotes the generated conformer. Then we can get RMSD $\Delta \;=\; \hat{\mathbf{x}_1}^\star- \mathbf{x}_1 $ by Kabsch alignment \citep{kabsch1976solution} where $\hat{\mathbf{x}_1}^\star$ is the conformer after alignment. This error can be considered as a kind of noise $ \Delta \sim \mathcal{N}(0,\sigma^\star\mathbf{I}_{3N})$ with an unknown real noise scale $\sigma^\star$. 

This suggests a simple shift: instead of starting from pure noise, we initialize sampling from upstream-generated conformers $\hat{\mathbf{x}_1}^\star$, thereby skipping the inherently hard-to-learn high-noise phase. Moreover, flow matching (FM) admits an arbitrary choice of the base distribution for $\mathbf{x}_0$ \citep{fm_pooladian2023multisample}.
Doing so alleviates early-stage error propagation and yields a smaller initial sampling error; it finally improves overall stability. Accordingly, we propose a refiner that further polishes conformers generated by the prior approaches rather than regenerating them from scratch. Our design addresses three challenges:
\begin{itemize}
\item[(a)] \textbf{Unknown scale.} With $\sigma^\star$ unknown at test time, how can we define an effective scale on the fly?
\item[(b)] \textbf{Scale realignment.} How should we set or adapt $\sigma$ to bypass the pure-noise phase and enter a well-trained regime?
\item[(c)] \textbf{Diversity preservation.} How can we improve quality without collapsing the diversity of upstream conformers?
\end{itemize}

\subsection{Refiner Definition}
\label{sec: method def-refiner}
\textbf{Base distribution $x_0$ in training}.
Since our goal is to \emph{refine} upstream conformers rather than regenerate them, we depart from the usual pure-noise initialization and adopt a base distribution $\mathbf{x_0}$ that retains the ground-truth signal. Therefore, we define
\begin{equation}
\label{eq: base distribution}
\mathbf{x}_0 \;=\; \mathbf{x}_1 \,+\, \sigma\,\varepsilon,
\qquad \varepsilon \sim \mathcal{N}(0,\mathbf{I}_{3N}).
\end{equation}

This is permitted by flow matching, which allows an arbitrary choice \citep{fm_pooladian2023multisample}. In implementation, we set $\sigma=1$, which has the following properties:

(1) At $t=0$, sampling does not start from pure noise but from a data-centered state that already contains signal, thereby skipping the pure-noise phase. Moreover, the scale $\sigma=1$ is well below the variance-exploding (VE) regime \citep{diff_song2020score}, which covers the data only under very large noise.

(2) The initial perturbation $\sigma \,\varepsilon$ exceeds the error level of prior methods. By introducing the Wilson--Hilferty approximation \citep{theory_Wilson–Hilferty}, it gives: when $\sigma=1$, for a conformer with $N=10$ heavy atoms perturbed by Gaussian noise $\sigma \varepsilon$, the upper RMSD bound with 95\% Confidence Interval is about $1.98 ~\text{\AA}$, which is notably larger than the typical errors of prior methods (details in Appendix~\ref{app: rmsd bound}). 
Because the schedule scales the noise as $(1-t)\sigma$, whose continuity in $t$ ensures value-range coverage of $t^{\star}\in(0,1)$: there exists $t^{\star}\in(0,1)$. 
Here, we denote $t^\star \in [0,1]$ as the unknown effective time at which the refiner’s noise scale matches that of the upstream conformer, i.e., $(1-t^\star)\,\sigma = \sigma^\star$.
such that
\begin{equation}
    (1-t^{\star})\,\sigma =\; \sigma^{\star},
    \qquad 
    t^{\star} \;=\; 1-\frac{\sigma^{\star}}{\sigma}.
\end{equation}
This justifies a sampling-time self-calibration: once $(1-t)\sigma \approx \sigma^{\star}$, the effective refinement begins at $t^{\star}$.

\textbf{Interpolant $\mathbf x_t$ and velocity $\mathbf u_t$ of the refiner.}
Given $\mathbf x_1$ and a base distribution defined by~\ref{eq: base distribution}.
We use the linear interpolant:
\begin{equation}
\label{eq:interpolant}
\begin{split}
\mathbf x_t
&= (1-t)\,\mathbf x_0 + t\,\mathbf x_1 + s(t)\,\mathbf z\\
&= \mathbf x_1 + (1-t)\,\sigma\varepsilon + s(t)\,\mathbf z .
\end{split}
\end{equation}
The corresponding velocity is
\begin{equation}
\label{eq: velocity define}
\mathbf u_t \;=\; \tfrac{\mathrm d}{\mathrm dt}\mathbf x_t
= -\,\sigma\varepsilon \;+\; s'(t)\,\mathbf z .
\end{equation}
Following \citet{baseline_hassan2024flow}, we schedule $s(t)=\sqrt{t(1-t)}$ to control the instantaneous velocity, which preserves value-range coverage of the noise scale:
\begin{equation}
\label{eq: stochastic}
s'(t)=\,\frac{1-2t}{2\sqrt{t(1-t)}}\mathbf z, 
\qquad
\mathbf u_t = -\, \sigma \varepsilon + \frac{1-2t}{2\sqrt{t(1-t)}}\,\mathbf z
\end{equation}
\textbf{Objective function}
We train a time-conditioned vector field $\mathbf u_\theta(\cdot,t)$ to match the target velocity along the interpolant:
\begin{equation}
\label{eq: loss}
\min_{\theta}\ \mathbb E\!\left[\left\|\,\mathbf u_\theta(\mathbf x_t,t, \mathcal G)\;-\;\mathbf u_t\,\right\|_2^2\right],
\qquad t\sim\mathrm{Unif}[0,1]
\end{equation}
\textbf{Sampling}, given an upstream sample, we set $\mathbf x_0=\widetilde{\mathbf x}$ and sampling
\begin{equation}
\label{eq: refiner-ode}
\frac{\mathrm d}{\mathrm dt}\mathbf x_t \;=\; \mathbf u_\theta(\mathbf x_t,t),
\qquad \mathbf x_{t=0}=\mathbf x_0,
\end{equation}
to obtain the refined conformer $\mathbf x_{t=1}$. The Detailed can be found in Algorithm~\ref{alg1}. We are also following the same correction strategy as ~\citet{baseline_hassan2024flow}

\paragraph{Design implications.}
The proposed base distribution and $t$-schedule address challenges \textbf{(a)} and \textbf{(b)} as follows: for \textbf{(a)}, by ensuring a smaller noise in the value range coverage. The conformers by the upstream model with error no larger than $\sigma\varepsilon$ lie within the refiner’s reachable range, enabling on-the-fly realignment without knowing $\sigma^\star$.
For \textbf{(b)}, the base distribution in Eq.~\ref{eq: base distribution} is data-centered rather than pure noise, so sampling does not begin in the pure-noise regime and thus bypasses it.

Also, this design introduces a new challenge:
\textbf{(d)} step time mismatch. Because effective refinement begins near $t^\star$ rather than at the real start of the sampling, the pair $(\mathbf{x}_t, t)$ may be distributionally mismatched, which can harm the quality. In the next section, we address challenges \textbf{(c)} and \textbf{(d)} by detailing the model’s representations and update rules, which maintain diversity and are robust to the $(\mathbf{x}_t, t)$ mismatch.

\begin{algorithm}[t]
\caption{\textsc{Refiner} (ODE)}
\label{alg1}
\begin{minipage}[t]{0.48\textwidth}
\textbf{Training}
\begin{algorithmic}[1]
\State \textbf{repeat}
\State \quad \textbf{sample}\; $\mathbf x_1,\;
\mathcal{G}=(\mathcal{V},\mathcal{E})\sim p_{\text{data}}$
\Statex \hspace{\algorithmicindent} $\varepsilon \sim \mathcal{N}(\mathbf 0,I_{3N})$, \;\; $t \sim \mathrm{Uniform}(0,1)$
\State \quad $\mathbf x_0 \gets \mathbf x_1 + \sigma\,\varepsilon$
\State \quad $\mathbf x_0 \gets \mathrm{Align}_{\text{Kabsch}}(\mathbf x_0,\,\mathbf x_1)$
\State \quad $\mathbf x_t \gets (1-t)\,\mathbf x_0 + t\,\mathbf x_1$
\State \quad $\mathbf u_t \gets \mathbf x_1 - \mathbf x_0$
\State \quad \textbf{Predict}\; $\hat{\mathbf u} \gets \mathbf u_\theta (\mathbf x_t, t, \mathcal{G})$
\State \quad $\mathcal{L} \gets \|\hat{\mathbf u} - \mathbf u_t\|_2^2$; \; update $\theta \leftarrow \theta - \eta \nabla_\theta \mathcal{L}$
\State \textbf{until} convergence
\end{algorithmic}
\end{minipage}\hfill
\begin{minipage}[t]{0.48\textwidth}
\textbf{Sampling}
\begin{algorithmic}[1]
\Require Generated sample $\hat{\mathbf x}$ and its graph $\mathcal{G}$
\Statex \hspace{\algorithmicindent} Trained refiner model $\mathbf u_\theta$
\Statex \hspace{\algorithmicindent} Number of steps $N$
\State Schedule $\{t_n\}_{n=0}^{N}$ with $t_0=0$, $t_N=1$
\State $\mathbf x \gets \hat{\mathbf x}$
\For{$n=0$ \textbf{to} $N-1$}
  \State $\Delta t_n \gets t_{n+1}-t_n$
  \State $\mathbf x \gets \mathbf x + \Delta t_n \cdot \mathbf u_\theta(\mathbf x, t_n, \mathcal{G})$
\EndFor
\State \Return $\mathbf x$
\end{algorithmic}
\end{minipage}
\end{algorithm}
\subsection{Properties}
\label{sec: prop}
Refiner behavior under a mismatch in time ($t^\star \neq t$), aiming to preserve diversity and ensure robustness so that already good states are not downgraded. Our analysis centers on how atom coordinates are represented by the dynamic part of the representation.

\textbf{Representation properties.}
We first analyze the model’s representation properties. Following prior denoising-based approaches \citep{baseline_xu2022geodiff, baseline_jing2022torsional, baseline_hassan2024flow}, we parameterize the refiner velocity model $\mathbf u_{\theta}(\mathbf x_t, t, \mathcal G)$ as in Eq.~\ref{eq: loss}. After removing components that are not dependent on $t$, including $\mathcal G$ or $\mathbf{x_1}$ (in Eq.~\ref{eq:interpolant}, the $\mathbf{x_1}$ is part of $\mathbf{x_t}$) , the remaining $t$-dependent representation can be written as 
$(1-t)\,\sigma\,\varepsilon \;+\; s(t)\,\mathbf z .$
Consequently, the magnitude of the velocity target defined in Eq.~\ref{eq: velocity define} is positively correlated with it.

\textbf{Relative representation.}
By SE(3)-equivariance, we may fix atom $i$ at the origin through a global rigid transform.
The representation is the collection of relative vectors $r_{ij}(t) \;=\; \mathbf{x}_j(t) - \mathbf{x}_i(t)$.
together with the neighbor set induced by a radius threshold $R$: $ \mathcal{D}_R(i,t) \;=\; \big\{\, j \neq i \;\big|\; \|r_{ij}(t)\| \le R \,\big\}.$
After alignment at $t=1$, denote the decomposition
\begin{equation}
\label{eq:decomp}
    r_{ij}(t) \;=\; r_{ij}(1) \,+\, \Delta r_{ij}(t),
\end{equation}
where $r_{ij}(1)$ is the static reference part and $\Delta r_{ij}(t)$ collects all time-varying perturbations.
Thus, the representation that may mismatch with $t$ includes: (i)  distance perturbation: $\Delta d_{ij}(t) $; (ii) angular deviation: $\overrightarrow{r_{ij}}(t)'$, (iii)  neighbor degree under radius $R$: $\mathcal{D}_R(i,t)$.

\begin{figure*}
  \centering
  \includegraphics[width=\linewidth,clip,trim=0 345 225 75]{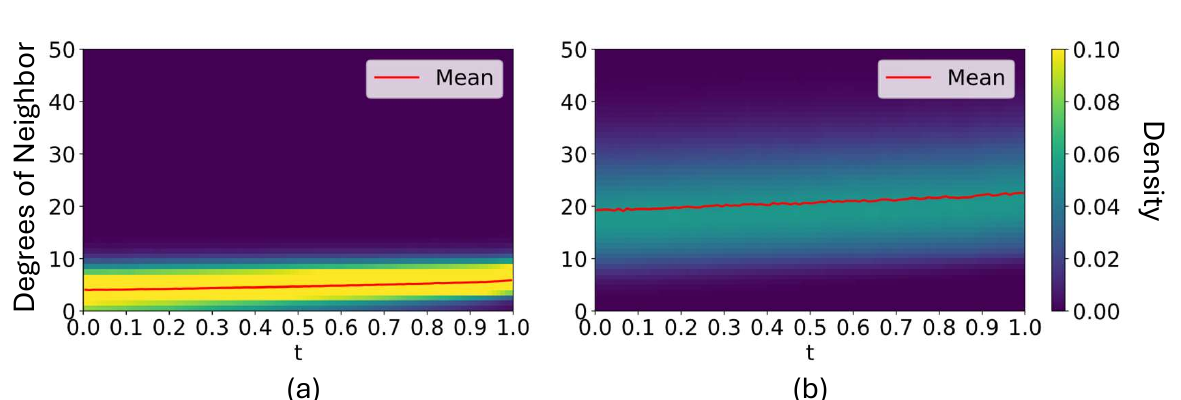}
   \caption{Comparison of neighbor degree distributions during training: (a) with a maximum radius of 2.5; (b) with a maximum radius of 5.0.}
  \label{fig: degree-of-neighbor}
\end{figure*}

\textbf{Representation properties when $t$ mismatch..}
We focus on scenarios with a time mismatch $t^\star > t$, where $(1-t^\star)\sigma < (1-t)\sigma$ and thus $\sigma^\star = (1-t^\star)\sigma < \sigma$.
We analyze the consequences of this mismatch for the position vector $r_{ij}$ and the neighbor degree $\mathcal{D}_R(i,t)$ as follows:

Firstly, we approximate the time-mismatch effect on a pair by an origin-centered Gaussian perturbation:
\begin{equation}
\label{eq:atom-perturb}
    r_{ij}(t) \;=\; r_{ij}(1) \;+\; \sigma_{\text{eff}}\,\varepsilon,
    \qquad
    \varepsilon \sim \mathcal N\!\big(0, I_3\big),
\end{equation}
with $\sigma_{\text{eff}}=\sqrt{2}$ when each endpoint has isotropic noise at both endpoints; equivalently, after fixing $x_i$ at the origin, the effective perturbation on $r_{ij}$ has variance $\sigma_i^2+\sigma_j^2$, which is equal to $2\sigma^2$ under identical scales.

Then we can get (i) when distance \& direction under time mismatch:
Both the distance perturbation $\,\Delta d(t)\,$ and the angular deviation of $\,r_{ij}(t)'\,$ relative to $\,r_{ij}\,$ are positively correlated with the noise scale $\,\sigma_{\text{eff}}\,$~\citep{anderson1958introduction,mardia2009directional}, therefore, configurations attainable under a smaller scale $\sigma ^\star$ are contained within those under a larger scale $\sigma$ .

(ii) when the neighbor degree under time mismatch:
The neighborhood size $|\mathcal D(i)|$ (with a fixed radius cutoff $R$) directly scales the magnitude of aggregated vector features.
Because neighbors at different distances encode distinct representations, we empirically quantify the degree by randomly subsampling under a fixed $R$ and counting $|\mathcal D(i)|$ (see Fig.~\ref{fig: degree-of-neighbor}).
As $t$ increases during training while the effective noise $(1-t)\sigma$ decreases, $|\mathcal D(i)|$ monotonically increases.
When test states correspond to $t^\star > t$, higher-degree configurations may be underrepresented relative to earlier-$t$ training samples, which can induce errors at the beginning of refinement. 

\textbf{Sampling dynamic properties when $t$ mismatch.}
We summarize properties relevant to the mismatch between $(\mathbf x_t,t)$ and $(\mathbf x_{t^\star},t^\star)$, assuming that noise drives the dynamics and correlates with the target.

1. Distance-dominant scaling under $\mathrm{SE}(3)$.
With an $\mathrm{SE}(3)$-equivariant backbone where orientation is not modeled explicitly, direction provides a limited independent signal. Update magnitudes scale primarily with distance perturbations, which are strongly correlated with the effective noise scale $\sigma_{\mathrm{eff}}$. 
Consequently, cleaner inputs with larger $t^\star$ induce smaller predicted velocities $\|\mathbf u_t\|$.

2. Degree mismatch co-occurs with low velocity.
Neighbor degree mismatch is most likely at higher noise. Mismatch is more likely to happen when the conformers generated by the upstream model are cleaner than the refiner’s current state, i.e., $t^\star>t$. Under this condition, the velocity will be relative low as well.

\textbf{Representation mismatch implications} Taken together, the dynamic properties of our representation directly address the above challenges:
\textbf{(c)} If a generated conformer is already near a target basin, the predicted velocities are low-magnitude, which ensures a small update and avoids basin switching, therefore preserving diversity.
\textbf{(d)} mismatch arises when the refiner’s current time step is smaller, i.e., $t^\star>t$.  This makes a warm-up phase in which some of the atoms' neighborhood degrees may be out of distribution, but the associated velocities are small, so errors won't be catastrophic. 

\section{Experimental Result}
\label{sec: experiment}
\begin{table}[t]
\caption{Performance of generated conformer ensembles on the GEOM-DRUGS test set, reported as Coverage (COV, \%) and Average Minimum RMSD (AMR, \AA). Coverage (COV) is computed with threshold $\delta=0.75~\text{\AA}$. The best performance is \textbf{bold}.}
\label{tab:geom-DRUGS}
\centering
\begin{tabular}{l|cccc|cccc} \toprule
                & \multicolumn{4}{c|}{Recall} & \multicolumn{4}{c}{Precision}  \\
                  & \multicolumn{2}{c}{COV $\uparrow$} & \multicolumn{2}{c|}{AMR $\downarrow$} & \multicolumn{2}{c}{COV  $\uparrow$} & \multicolumn{2}{c}{AMR $\downarrow$} \\
Method & Mean & Med & Mean & Med & Mean & Med & Mean & Med \\
\midrule
GeoMol                      & 44.60 & 41.40 & 0.875 & 0.834 & 43.00 & 36.40 & 0.928 & 0.841 \\
GeoDiff                     & 42.10 & 37.80 & 0.835 & 0.809 & 24.90 & 14.50 & 1.136 & 1.090 \\ 
Torsional Diff.             & 72.70 & 80.00 & 0.582 & 0.565 & 55.20 & 56.90 & 0.778 & 0.729 \\ 
MCF-L (242M)                 & 85.10 & 92.86 & 0.390 & 0.343 & 66.63 & 70.00 & 0.623 & 0.546 \\ 
ET-Flow (8.3M)               & 80.15 & 85.71 & 0.458 & 0.429 & 73.89 & 80.56 & 0.556 & 0.494 \\ 
DMT-L (150M)                 & 85.95 & 91.98 & 0.378 & 0.353 & 67.97 & 71.97 & 0.599 & 0.529 \\ 
\midrule
MCF-L + \textbf{Refiner} (8.3M)     & 86.44 & 93.68 & 0.368 & 0.330 
                                    & 72.07 & 78.41 & 0.550 & 0.480 \\ 
ET-Flow  + \textbf{Refiner}         & 80.29 & 85.11 & 0.439 & 0.410 
                                    & 74.58 & \textbf{81.59} & 0.530 & 0.467 \\ 
DMT-L + \textbf{Refiner}            & \textbf{87.47} & \textbf{94.12} & \textbf{0.349} & \textbf{0.319}
                                    & \textbf{75.91} & 81.51 & \textbf{0.497} & \textbf{0.446} \\ 
\midrule
Boost (\%)    &1.95	&1.97	&7.44	&8.36	&2.73	&1.28	&12.03	&10.94 \\
\bottomrule
\end{tabular}
\end{table}

\begin{table}[t]
\caption{Performance of generated conformer ensembles on the GEOM-QM9 test set, reported as Coverage (COV, \%) and Average Minimum RMSD (AMR, \AA). Since recent work already achieves 100\% median COV at the commonly used threshold $\delta=0.5$~\AA\ and a median AMR below $0.05$~\AA, we adopt the more challenging COV threshold of $\delta=\textbf{0.05}$~\AA. The best performance is \textbf{bold}.}
\label{tab:geom-QM9}
\centering
\begin{tabular}{l|cccc|cccc} \toprule
                & \multicolumn{4}{c|}{Recall} & \multicolumn{4}{c}{Precision}  \\
                  & \multicolumn{2}{c}{COV (\textbf{0.05}) $\uparrow$} & \multicolumn{2}{c|}{AMR $\downarrow$} & \multicolumn{2}{c}{COV  (\textbf{0.05}) $\uparrow$} & \multicolumn{2}{c}{AMR $\downarrow$} \\
Method & Mean & Med & Mean & Med & Mean & Med & Mean & Med \\
\midrule
GeoDiff                     & - & - & 0.297 & 0.229 & - & - & 1.524 & 0.510 \\
GeoMol                      & - & - & 0.225 & 0.193 & - & - & 0.270 & 0.241   \\ 
Torsional Diff.             & - & - & 0.178 & 0.147 & - & - & 0.221 & 0.195   \\ 
MCF-B (62M)                 & 66.82 & 67.86 & 0.101 & 0.050 & 61.18 & 64.29 & 0.117 & 0.059 \\ 
ET-Flow (8.3M)              & 75.72 & 87.23 & 0.083 & 0.031 & 70.32 & 75.00 & 0.114 & 0.053 \\ 
DMT (55M)                   & 72.90 & 83.33 & 0.087 & 0.036 & 67.76 & 75.00 & 0.107 & 0.047 \\ 
\midrule
MCF-B + \textbf{Refiner} (8.3M)     & 74.87 & 81.82 & 0.100 & 0.035
                                    & 77.41 & 94.44 & 0.101 & 0.023 \\ 
ET-Flow + \textbf{Refiner}          & 78.40 & 88.89 & 0.076 & 0.028
                                    & 77.36 & 89.94 & 0.103 & 0.031 \\ 
DMT-B + \textbf{Refiner}            & \textbf{79.50} & \textbf{89.44} & \textbf{0.070} & \textbf{0.026} 
                                    & \textbf{80.37} & \textbf{97.92} & \textbf{0.076} & \textbf{0.021} \\ 
\midrule
Boost (\%)    
&3.86	&2.22	&11.22	&14.60	&13.72	&22.78	&35.77	&55.32\\

\bottomrule
\end{tabular}
\end{table}

We evaluate the generator–refiner pipeline and, via controlled studies, isolate the refiner’s contribution to empirically validate our theoretical analysis by following the research questions:

\textbf{RQ1: Effectiveness.} Compared with generator-only sampling, does the generator–refiner pipeline produce higher-quality conformers, preserve diversity, and do so with fewer steps? (Sections~\ref{sec: exp geo} and~\ref{sec: exp chem})

\textbf{RQ2: Refiner impact.} For upstream-generated conformers, what proportion are \emph{improved} versus \emph{downgraded} by the refiner? (Section~\ref{sec: exp refiner})

\textbf{RQ3: Sampling dynamics.} What dynamics does the refiner exhibit, and do these dynamics align with our theoretical analysis (as assessed by empirical fits)? (Section~\ref{sec:exp dynamic})

RQ1–RQ2 focus on performance comparisons, whereas RQ3 examines sampling dynamics and associated property behavior.

\subsection{Setup}
\paragraph{Dataset.} We evaluate on the GEOM dataset \citep{axelrod2022geom}.
GEOM-DRUGS is the largest relevant subset (304k drug-like molecules). 
We also train and evaluate on GEOM-QM9, a more established benchmark with smaller molecules. We follow ~\citet{baseline_ganea2021geomol} random splits of 80\%/10\%/10\% into train/validation/test. Following~\citet{baseline_ganea2021geomol, baseline_jing2022torsional}, we use the same 1,000 random test molecules from the test set. The dataset splits are 106,586/13,323/1,000 (GEOM-QM9) and 243,473/30,433/1,000 (GEOM-DRUGS) molecules.

\paragraph{Implementation Detail and Baseline.}
We implement our refiner by fine-tuning the open-source ET-Flow \citep{baseline_hassan2024flow} architecture and weights. For fine-tuning, we reduce the learning rate from 0.007 to 0.001 and adopt a Cosine-Warmup learning rate schedule; otherwise, architectural choices and hyperparameters follow ET-Flow. At the sampling process, the refiner serves as a post-processor on conformers produced by the three most recent conformer generative models based on denoising: MCF \citep{baseline_wang2024swallowing}, ET-Flow \citep{baseline_hassan2024flow}, and DMT \citep{baseline_liu2025next}.
For upstream sampling, we use the official open source code implementations and trained weights. We compare the final performance with recent advanced models \citep{baseline_ganea2021geomol, baseline_xu2022geodiff, baseline_jing2022torsional, baseline_wang2024swallowing, baseline_hassan2024flow, baseline_liu2025next}. 

\paragraph{Evaluation Metric.} 
Following \citet{baseline_ganea2021geomol, baseline_jing2022torsional}, for each molecule with \(K\) reference conformers, we generate \(2K\) candidate conformers.
We report Average Minimum RMSD (AMR)-precision (quality), AMR–recall (diversity), and Coverage (COV) (see the Appendix \ref{app: geometric} for details).
Following  \citet{baseline_jing2022torsional}, we also evaluate chemical similarity using properties computed with xTB \citep{gfnxtb}: total energy \(E\), dipole moment \(\mu\), HOMO--LUMO gap \(\Delta\epsilon\), and minimum energy \(E_{\min}\).
Finally, because our refiner aims to improve conformers, we additionally report improvement and downgrade rates relative to each conformer's baseline quality.

\subsection{Ensemble RMSD and Sampling Efficiency}
\label{sec: exp geo}
To demonstrate higher quality with fewer sampling steps, we adopt a stricter budget: baselines use a single generator with 50 sampling steps, while our pipeline (generator plus refiner) uses 40 steps in total (20 for generation and 20 for refinement). Following the original papers, we run MCF and DMT at their largest-scale configurations: Large (L) on GEOM-DRUGS and Basic (B) on GEOM-QM9. 

As shown in Table~\ref{tab:geom-DRUGS} and Table~\ref{tab:geom-QM9}, our method surpasses baselines. On the precision metric (AMR), the median on GEOM-DRUGS decreases by 10.94\%, and on GEOM-QM9, the error is roughly halved. On the diversity metric (recall), we see a smaller improvement: 8.36\% on GEOM-DRUGS AMR-median and 14.6\% on GEOM-QM9 AMR-median. These recall gains are driven mainly by improving conformer quality: for each ground truth conformer, the generated conformer that was previously its nearest match remains the match but fits better after refinement. Thus, the refiner substantially improves quality and preserves diversity. See the Appendix~\ref{app ablation} for additional ablations on the number of steps.

\begin{wraptable}{r}{0.5\columnwidth}
  \vspace{-\intextsep}
  \centering
  \footnotesize
  \caption{Median Boltzmann-weighted errors of ensemble properties between sampled and generated conformers: $E$, $\Delta \varepsilon$ in kcal/mol, and $\mu$ in debye; median and $E_{min}$ in kcal/mol.}
  \label{tab:chem}
  \begin{tabular}{l|cccc}
    \toprule
    & $E$ & $\mu$ & $\Delta \epsilon$ & $E_{\text{min}}$ \\
    \midrule
GeoDiff	                    & 0.31	& 0.35	& 0.89	& 0.39 \\
GeoMol	                    & 0.42	& 0.34	& 0.59	& 0.40 \\
Torsional Diff.	            & 0.22	& 0.35	& 0.54	& 0.13 \\
MCF-L	                    & 0.68	& 0.28	& 0.63	& 0.04 \\
ET-Flow	                    & 0.23	& 0.19	& \textbf{0.38}	& 0.02 \\
\midrule
MCF-L + \textbf{Refiner}	& \textbf{0.20}	& 0.23	& \textbf{0.38}	& 0.02  \\
ET-Flow + \textbf{Refiner}	& 0.21	& \textbf{0.18}	& 0.39	& \textbf{0.01} \\
    \bottomrule
  \end{tabular}
\end{wraptable}

\subsection{Chemical Property}
\label{sec: exp chem}
We also compare the chemical similarity between generated and ground truth conformers. We follow and use the same 100-molecule subset of \citet{baseline_jing2022torsional}. For a molecule with $K$ ground truth conformers, we randomly select $\min(2K, 32)$ generated conformers, relax them with GFN2-xTB\citep{gfnxtb}, and compare Boltzmann-weighted ensemble properties between the generated and ground truth sets. Using xTB\citep{gfnxtb}, we compute energy ($E$), dipole moment ($\mu$), HOMO–LUMO gap ($\Delta\varepsilon$), and minimum energy ($E_{\min}$). Table~\ref{tab:chem} reports median errors, showing that our method can get better chemically accurate ensembles. 
\subsection{Improvement and Downgrade Rate on RMSD Ensemble}
\label{sec: exp refiner}
Beyond the macro-level improvement in Section~\ref{sec: exp geo}, we test at a micro level whether the refiner improves conformers for different upstream generative models. For every conformer, we compute precision RMSD before and after refinement and apply multi-tolerance thresholds $\tau$ (\AA) to label outcomes as improvement or downgrade. Each conformer is paired only with its own refined counterpart (one-to-one). We then report improvement/downgrade rates for each group in Tables~\ref{tab: imp-downgrade DRUGS} and~\ref{tab: imp-downgrade QM9}. Across all comparisons and thresholds, our refinement yields improvement rates that are multiples of the downgrade rates.

\begin{table}[t]
\centering
\caption{Improvement rate (IR) / Downgrade rates (DR) (\%) using RMSD Precision with a  relative tolerance $\tau$ (\AA) of GEOM-Drugs}
\label{tab: imp-downgrade DRUGS}
\begin{tabular}{l|cc|cc|cc}
\toprule
$\tau$
& \multicolumn{2}{|c}{MCF $\rightarrow$  + Refiner}
& \multicolumn{2}{|c}{ET-Flow $\rightarrow$ + Refiner}
& \multicolumn{2}{|c}{DMT $\rightarrow$  + Refiner} \\
\cmidrule(lr){2-3}\cmidrule(lr){4-5}\cmidrule(lr){6-7}
& IR & DR & IR & DR & IR & DR \\
\midrule
  0.05 & 37.5 & 16.7 & 35.3 & 5.6 &  22.7 & 7.5 \\
  0.10 & 23.9 & 7.0  & 15.7 & 1.2 &  9.7 &  1.8 \\
  0.20 & 9.6  & 1.4  & 2.6  & 0.1 &  2.1 &  0.2 \\
  0.50 & 0.5  & 0.0  & 0.0  & 0.0 &  0.0 &  0.0  \\
\bottomrule
\end{tabular}
\end{table}

\begin{figure*}
  \centering
  \includegraphics[width=\linewidth,clip,trim=0 420 150 85]{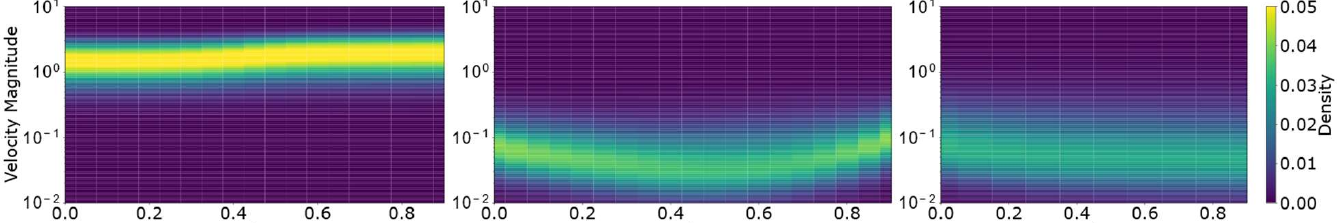}
  \caption{Velocity fields on GEOM–QM9: (a) ET–Flow sampling; (b) Refiner; (c) refiner with randomized $t$.}
  \label{fig: velocity}
\end{figure*}

\subsection{Empirical examination of sampling dynamics}
\label{sec:exp dynamic}
We empirically examine the sampling dynamics induced by the properties introduced in Section~\ref{sec: prop}, which drive the refiner.
To avoid architectural side effects, the upstream is fixed to ET-Flow \citep{baseline_hassan2024flow}, which is the refiner fine-tuning on.
We use GEOM–QM9, whose stronger upstream quality implies a larger $(\mathbf{x}_t,t)$ vs.\ $(\mathbf{x}_{t^\star},t^\star)$ mismatch, making it a stricter test.

\begin{wrapfigure}{r}{0.5\columnwidth}
  \vspace{-\intextsep} 
  \centering
  \includegraphics[width=\linewidth,clip,trim=0 240 290 100]{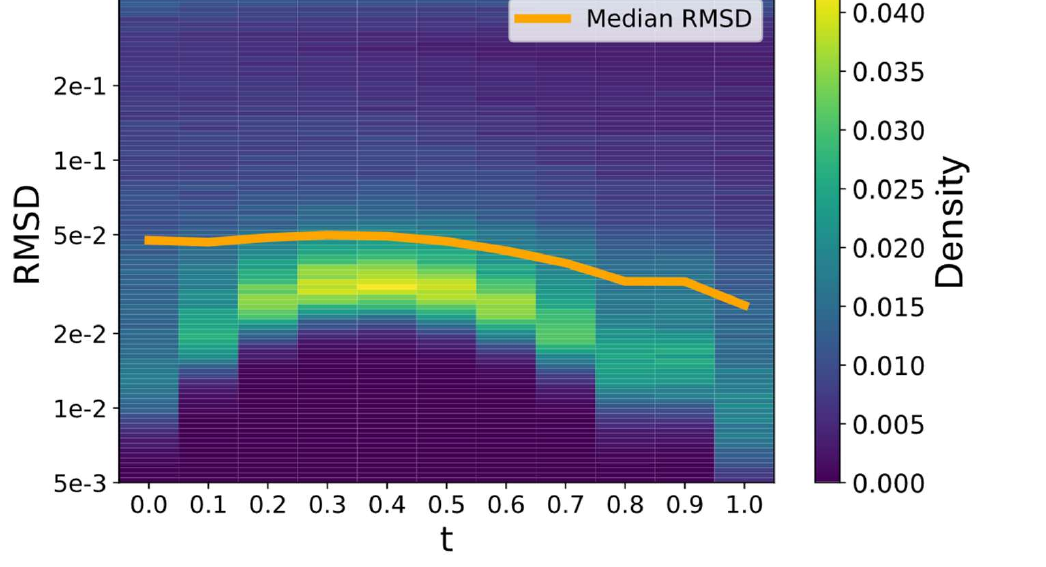}
  \caption{GEOM-QM9 AMR–precision dynamics during refinement}
  \label{fig: refiner-dynamic}
\end{wrapfigure}

\textbf{Self-calibration and two stages.}
Fig.~\ref{fig: refiner-dynamic} traces the RMSD throughout refinement. A short \emph{warm-up} period ($t\in[0,0.3]$) exhibits a small RMSD increase due to neighbor-degree mismatch; then, once the neighbor degrees are aligned better under the $t$-schedule, RMSD decreases monotonically during \emph{refinement}. At $t=1$, the median precision AMR declines from $\approx0.05$~(\AA) to $\approx0.03$~(\AA), with a noticeable subset attaining RMSD $<0.01$~(\AA), which is rarely observed with a single flow-matching model.

\textbf{Velocity dynamics.}
The second key property is the velocity behavior. In Fig.~\ref{fig: velocity}, (a) ET-Flow shows typical speeds exceeding 1; (b) the refiner concentrates below $\sim\!0.1$; and (c) when we randomize the input $t$ to the refiner, the low-velocity pattern persists. These observations indicate: when the upstream conformer starts in a relatively low-noise state, even under time-step mismatch, the refiner produces low velocities that avoid catastrophic errors during the warm-up stage. Also, compared with the much higher velocity during the sampling of the generation process, such low velocities by the refiner can keep atoms within their current basins and thereby preserve diversity.

\section{Conclusion}
In this paper, we propose a flow-matching based refiner for molecular conformer generation. 
At sampling time, the refiner samples directly on conformers generated by diverse upstream denoising models, re-aligning the perturbed conformers on the fly without requiring access to the upstream model.
By rescheduling the noise scale, the method bypasses the ill-trained low-SNR regime and early-step error propagation, yielding a clear second-stage quality gain. 
Empirically, with fewer total steps, our denoising model + refiner pipeline achieves better performance, and the improvement rate substantially exceeds the downgrade rate.
\clearpage

\bibliography{reference}
\bibliographystyle{conference}

\clearpage
\appendix
\begin{table}[ht]
\caption{Performance of generated conformer ensembles on the GEOM-DRUGS test set, reported as Coverage (COV, \%) and Average Minimum RMSD (AMR, \AA). Coverage (COV) is computed with threshold $\delta=0.75~\text{\AA}$. The best performance is \textbf{bold}.} 
\label{tab:geom-DRUGS-detail}
\centering
\begin{tabular}{l|c|cccc|cccc} \toprule
              &  & \multicolumn{4}{c|}{Recall} & \multicolumn{4}{c}{Precision}  \\
               &   & \multicolumn{2}{c}{COV $\uparrow$} & \multicolumn{2}{c|}{AMR $\downarrow$} & \multicolumn{2}{c}{COV  $\uparrow$} & \multicolumn{2}{c}{AMR $\downarrow$} \\
Method & steps & Mean & Med & Mean & Med & Mean & Med & Mean & Med \\
\midrule
MCF-L	&10	&84.52	&92.86	&0.444	&0.408	&65.63	&68.75	&0.668	&0.608 \\
MCF-L	&20	&85.82	&93.18	&0.400	&0.369	&66.43	&69.54	&0.637	&0.560 \\
MCF-L	&50	&85.10	&92.86	&0.390	&0.343	&66.63	&70.00	&0.623	&0.546 \\
MCF-L	&100&85.06	&92.12	&0.388	&0.348	&66.70	&70.45	&0.621	&0.539 \\
\midrule
MCF-L + Refiner	&10 + 10	&85.61	&93.67	&0.386	&0.348	
                            &71.82	&76.67	&0.559 &	0.493 \\
MCF-L + Refiner	&20 + 20	&\textbf{86.44}	&\textbf{93.68}	&\textbf{0.368}	&\textbf{0.330}	
                            &\textbf{72.07}	&\textbf{78.41}	&\textbf{0.550} &\textbf{0.480} \\

\midrule
\midrule
DMT-L	&10	&85.82	&91.99	&0.391	&0.364	&66.62	&70.50	&0.625	&0.564 \\
DMT-L	&20	&85.58	&92.44	&0.384	&0.358	&67.48	&71.65	&0.608	&0.532 \\
DMT-L	&50	&85.95	&91.98	&0.378	&0.353	&67.97	&71.97	&0.599	&0.529 \\
DMT-L	&100	&85.80	&92.30	&0.375	&0.346	&67.90	&72.50	&0.598	&0.527 \\
\midrule
DMT-L + Refiner	&10 + 10	&\textbf{87.66}	&93.45	&0.352	&0.329	
                            &74.92	&80.97	&0.512	&0.459 \\
DMT-L + Refiner	&20 + 20	&87.47	&\textbf{94.12}	&\textbf{0.349}	&\textbf{0.319}	
                            &\textbf{75.91}	&\textbf{81.51}	&\textbf{0.497}	&\textbf{0.446} \\

\midrule
\midrule
ET-Flow	&10	&79.41	&85.02	&0.467	&0.439	&72.33	&78.77	&0.577	&0.515 \\
ET-Flow	&20	&79.28	&84.74	&0.467	&0.437	&73.34	&80.00	&0.562	&0.499\\
ET-Flow	&50	&80.15	&85.71	&0.458	&0.429	&73.89	&80.56	&0.556	&0.494\\
ET-Flow	&100	&79.78	& 84.15	&0.462	&0.436	&73.70	&80.00	&0.561	&0.504\\
\midrule
ET-Flow + Refiner	&10 + 10	&80.20	&\textbf{85.71}	&0.445	&0.411	
                                &73.47	&81.25	&0.547	&0.477\\
ET-Flow + Refiner	&20 + 20	&\textbf{80.29}	&85.11	&\textbf{0.439}	&\textbf{0.410}	
                                        &\textbf{74.58}	&\textbf{81.59}	&\textbf{0.530}	&\textbf{0.467} \\

\bottomrule
\end{tabular}
\end{table}

\section{Appendix}
\subsection{RMSD upperbound with noise scale}
\label{app: rmsd bound}
After removing six $\mathrm{SE}(3)$ degrees of freedom, the non-rigid subspace has $d:=3N-6$ degrees of freedom. we approximate the error $\Delta$ as isotropic Gaussian noise in coordinate space with unknown scale $\sigma^\star$, we can get:
\begin{equation}
\label{eq:delta-chi}
\|\Delta\|^{2} \;\sim\; \sigma^{\star 2}\,\chi^{2}_{d},
\end{equation}
where $\chi^{2}_{d}$ is a chi-square random variable with $d$ degrees of freedom. 

Consequently,
\begin{equation}
\label{eq:rmsd-chi}
\mathrm{RMSD}(\widetilde{\mathbf x},\mathbf x_1)
\;=\; \sigma^\star\,\sqrt{\frac{1}{N}\chi^2_d}
\end{equation}

By the Wilson--Hilferty approximation \citep{theory_Wilson–Hilferty}, we have:
\begin{equation}
\label{eq:WH}
\left(\frac{X}{d}\right)^{\!1/3} \;\approx\; \mathcal N\!\left(1-\frac{2}{9d},\,\frac{2}{9d}\right)
\quad\text{for } X\sim\chi^2_d.
\end{equation}

by Eq.\ref{eq:rmsd-chi}, since $\mathrm{RMSD}^2 = (\sigma^{\star2}/N)\,X$ with $X\sim\chi^2_d$,
\begin{equation}
\label{eq:WH-RMSD}
\mathrm{RMSD}
\;\approx\; \sigma^\star\,
\sqrt{\frac{d}{N}\Big(1-\frac{2}{9d}+z\sqrt{\frac{2}{9d}}\Big)^{3}}\,,\qquad z\sim\mathcal N(0,1).
\end{equation}

Hence, denoting the standard-normal quantile at confidence level $k$ by $Q_k$, the RMSD quantile at scale $\sigma^\star$ is:
\begin{equation}
\label{eq:RMSD-bound}
\mathrm{RMSD}_{(Q_k,\sigma^\star)}\;\approx\;\sigma^\star\,
\sqrt{\frac{d}{N}\left(1-\frac{2}{9d}+Q_k\sqrt{\frac{2}{9d}}\right)^{3}}\,.
\end{equation}

\begin{table}[ht]
\caption{Performance of generated conformer ensembles on the GEOM-QM9 test set, reported as Coverage (COV, \%) and Average Minimum RMSD (AMR, \AA). Coverage (COV) is computed with threshold $\delta=0.05~\text{\AA}$. The best performance is \textbf{bold} }.
\label{tab:geom-Qm9-detail}
\centering
\begin{tabular}{l|c|cccc|cccc} \toprule
              &  & \multicolumn{4}{c|}{Recall} & \multicolumn{4}{c}{Precision}  \\
               &   & \multicolumn{2}{c}{COV $\uparrow$} & \multicolumn{2}{c|}{AMR $\downarrow$} & \multicolumn{2}{c}{COV  $\uparrow$} & \multicolumn{2}{c}{AMR $\downarrow$} \\
Method & steps & Mean & Med & Mean & Med & Mean & Med & Mean & Med \\
\midrule
\midrule
MCF-B	&10	&51.60 	&50.00 	&0.115	&0.067	&43.57 	&34.00 	&0.144	&0.094\\\
MCF-B	&20	&62.13 	&62.50 	&0.108	&0.056	&55.73 	&50.00 	&0.124	&0.068\\
MCF-B	&50	&66.83 	&67.86 	&0.101	&0.050	&61.18 	&64.29 	&0.117	&0.059\\
MCF-B	&100 &68.90 &75.00 	&0.099	&0.047	&64.00 	&69.12 	&0.112	&0.052\\
\midrule
MCF-B + Refine	&10+10	&74.32 	&80.00 	&\textbf{0.097}	&0.037	
                        &76.97 	&94.23 	&\textbf{0.100}	&0.025\\
MCF-B + Refine	&20+20	&\textbf{74.87} &\textbf{81.82} &0.101	&\textbf{0.035}	
                        &\textbf{77.41} &\textbf{94.44} &0.101	&\textbf{0.023}\\
\midrule
\midrule
DMT-B	&10	 &69.64 	&77.78 	&0.089	&0.041	&61.37 	&66.29 	&0.118	&0.062\\
DMT-B	&20	 &71.71 	&80.00 	&0.088	&0.038	&66.21 	&75.00 	&0.109	&0.048\\
DMT-B	&50	 &72.90 	&83.33 	&0.087	&0.036	&67.76 	&75.00 	&0.107	&0.047\\
DMT-B	&100 &73.75 	&83.33 	&0.085	&0.036	&68.39 	&75.00 	&0.103	&0.047\\
\midrule
DMT-B + Refine	&10+10	&\textbf{79.74} &88.89 	&\textbf{0.069}	&0.028	
                        &79.32 	&93.75 	&0.079	&0.025\\
DMT-B + Refine	&20+20	&79.50 	&\textbf{89.44} 	&0.070	&\textbf{0.026}	
                        &\textbf{80.37} 	&\textbf{97.92} 	&\textbf{0.076}	&\textbf{0.021}\\
\midrule
\midrule
ET-Flow	&10	    &75.07 	&87.50 	&0.082	&0.034	&67.49 	&73.07  &0.126	&0.066\\
ET-Flow	&20	    &75.65 	&85.71 	&0.083	&0.033	&69.40 	&75.00 	&0.121	&0.056\\
ET-Flow	&50	    &75.72 	&87.23 	&0.083	&0.031	&70.32 	&75.00 	&0.114	&0.053\\
ET-Flow	&100	&76.55 	&87.50 	&0.077	&0.030	&70.67 	&79.75  &0.115	&0.047\\
\midrule
ET-flow + Refine	&10+10	&\textbf{78.66} 	&\textbf{89.74} 	&\textbf{0.075}	&\textbf{0.027}	&76.73 	&87.50 	&0.106	&0.037\\
ET-flow + Refine	&20+20	&78.40 	&88.89 	&0.076	&0.028	&\textbf{77.36} 	&\textbf{89.94} 	&\textbf{0.103}	&\textbf{0.031}\\
\bottomrule
\end{tabular}
\end{table}

\begin{table}[t]
\centering
\caption{Improvement rate (IR) and Downgrade rate (DR) (\%) computed by RMSD-precision with a relative tolerance $\tau$ (\AA) on GEOM-QM9. Because the upstream error scale on QM9 is smaller than on GEOM-DRUGS, we adopt a tighter tolerance.}
\label{tab: imp-downgrade QM9}
\begin{tabular}{l|cc|cc|cc}
\toprule
$\tau$
& \multicolumn{2}{|c}{MCF $\rightarrow$  + Refiner}
& \multicolumn{2}{|c}{ET-Flow $\rightarrow$ + Refiner}
& \multicolumn{2}{|c}{DMT $\rightarrow$  + Refiner} \\
\cmidrule(lr){2-3}\cmidrule(lr){4-5}\cmidrule(lr){6-7}
& IR & DR & IR & DR & IR & DR \\
\midrule
  0.02  &  69.2 & 15.3 & 29.2 & 5.3 & 50.7 & 5.7 \\
  0.05  &  45.1 & 12.5 & 15.1 & 1.7 & 28.7 & 2.2 \\
  0.10  &  29.0 & 9.8  & 7.7  & 0.4 & 14.5 & 0.6 \\
  0.20  &  15.6 & 8.0  & 3.1  & 0.1 & 5.5  & 0.1 \\
\bottomrule
\end{tabular}
\end{table}

\subsection{Geometry Metric}
\label{app: geometric}
Following \citet{baseline_ganea2021geomol, baseline_jing2022torsional, baseline_hassan2024flow}, the following works have used the so-called Average Minimum RMSD (AMR) and Coverage (COV) for Precision(P): Quality and Recall(R): the diversity, measured when generating twice as many conformers as provided by CREST. For $K=2L$ let $\{C^*_l\}_{l \in [1, L]}$ for groundtruth and $\{C_k\}_{k \in [1, K]}$ for generated conformer.
\begin{equation}
\begin{aligned}
\text{COV-R} &:= \frac{1}{L}\, \bigg\lvert\{ l \in [1..L]: \exists k \in [1..K], RMSD(C_k, C^*_l) < \delta \, \bigg\rvert\\
\text{AMR-R} &:= \frac{1}{L} \sum_{l \in [1..L]}  \min_{ k\in [1..K]} RMSD(C_k, C^*_l)
\end{aligned}
\end{equation}
The $\delta$ is the coverage threshold, and the precision metrics are obtained by swapping ground truth and generated conformers.

\subsection{Additional Results: Refiner vs.\ More Sampling Steps}
\label{app ablation}
We present a detailed ablation comparing (i) increasing the upstream sampler’s steps and (ii) applying the refiner at equal or lower steps. Concretely, we evaluate base models at 10/20/50/100 steps and contrast them with “$k$ steps + refiner” ($10{+}10$, $20{+}20$). Results of Geom-Drugs can be found in Table~\ref {tab:geom-DRUGS-detail} and GEOM-QM9 can be found in Table~\ref{tab:geom-Qm9-detail}.

\subsection{Conformer-level improvement (QM9)}
Complementing Section~\ref{sec: exp refiner}, we report per-conformer improvement/downgrade on GEOM–QM9. For each conformer, we compute precision RMSD before/after refinement, apply multi-tolerance thresholds $\tau$ (\AA) to label \textbf{improvement} vs.\ \textbf{downgrade}. Details can be found in Table~\ref{tab: imp-downgrade QM9}.


\end{document}